\documentclass[11pt]{article}
\usepackage[utf8]{inputenc}
\usepackage[T1]{fontenc}
\usepackage{graphicx}
\usepackage{longtable}
\usepackage{wrapfig}
\usepackage{rotating}
\usepackage[normalem]{ulem}
\usepackage{amsmath}
\usepackage{amssymb}
\usepackage{capt-of}
\usepackage{hyperref}
\usepackage{authblk}
\usepackage{times}
\usepackage{soul}
\usepackage{todonotes}

\usepackage{enumitem}
\usepackage{comment}
\usepackage[a4paper, total={6.6in, 8in}]{geometry}
\author{Serge Sharoff, \small ORCID: 0000-0002-4877-0210}
\date{Published in \emph{Handbook of Democracy in the Era of Artificial Intelligence} edited by Evangelos Pournaras, Srijoni Majumdar, Carina Ines Hausladen, and Dirk Helbing. 2026.}
\title{The Almost Intelligent Revolution: Options for Scaling Up Deliberation and Empowering People with AI}
\hypersetup{
 pdfauthor={Serge Sharoff, \small ORCID: 0000-0002-4877-0210},
 pdftitle={The Almost Intelligent Revolution: Options for Scaling Up Deliberation and Empowering People with AI},
 pdfkeywords={},
 pdfsubject={},
 pdfcreator={Emacs 30.2 (Org mode 9.7.11)}, 
 pdflang={English}}
\begin{document}

\setlist{itemsep=5pt,parsep=0pt}

\affil{School of Languages, Cultures and Societies, University of Leeds, UK, \\E-mail: s.sharoff@leeds.ac.uk}
\maketitle

\begin{abstract}
The increasing prominence of Large Language Models (LLMs) in public discourse presents both opportunities and challenges for democratic deliberation. While red teaming strategies help mitigate specific risks, broader concerns persist regarding linguistic constraints, biases, and the sycophantic tendencies of LLMs.  This chapter explores how LLMs can be used to significantly scale up and democratise deliberation, particularly in fostering inclusivity and empowering traditionally marginalised groups. Drawing on concepts from Systemic-Functional Linguistics, the chapter examines how variations across language users (for example, with respect to socio-demographic groups) and across language use (for example, with respect to communicative functions) shape participation in AI-supported deliberation.  The chapter presents AI-driven deliberation studies and assesses their potential to scaffold argumentation, enhance access, and reduce the influence of exclusionary linguistic norms and biases which are embedded in prestigious registers.  At the same time, the chapter cautions against both overclaiming, which leads to unrealistic expectations, and underclaiming, which risks missed opportunities for AI-assisted engagement.  The chapter concludes by identifying future research directions to maximise the democratic potential of AI-assisted participation while embedding ethical safeguards to counteract the reproduction of linguistic inequalities.

\textbf{Keywords}: Large Language Models, communication, democratic deliberation, biases
\end{abstract}
\section{Introduction}
\label{sec:org6bc5ded}
Representative democracy is based on delegating policy matters to elected representatives, while participatory democratic processes aim at involving the stakeholders directly into decision making.  Participatory democratic processes emphasise deliberation and consensus-building, with their roots traced back to Jürgen Habermas \cite{bachtiger18deliberative}.  Modern democratic institutions push for greater involvement of the stakeholders to promote the participation of citizens and civil society organisations in policy making.  For example, the European Commision has recently established a Centre on Participatory and Deliberative Democracy.\footnote{\url{https://knowledge4policy.ec.europa.eu/participatory-democracy/}}

Mutual discussion and decision-making among a wide range of stakeholders is a worthwhile goal.  However, the process of deliberation runs through the medium of language, so that the stakeholders are required to have effective communication skills in order to understand others' arguments and express their own arguments convincingly.  Jürgen Habermas suggested the concept of the "ideal speech situation" as a set of conditions under which participants in a discourse can engage in rational communication free from distortions.  In particular, as one of these conditions, he mentions \emph{Equality}, when all participants have an equal opportunity to speak and be heard \cite{bachtiger18deliberative}.

The ideal speech situation provides a valuable framework on how communication should ideally occur.  However, the linguistic dimension of participation poses challenges for many stakeholders, because complex grammar and vocabulary as well as abstract reasoning can create barriers to their meaningful engagement.  Many social institutions develop their own communicative norms which often exhibit heavy use of specific technical jargon or culturally specific expressions \cite{halliday99}, while individuals with limited proficiency in using these norms may struggle to articulate their perspectives, fully grasp others' arguments, or respond effectively.  As a result, these linguistic constraints can lead to the exclusion of certain voices from deliberative processes, even the voices of the most relevant stakeholders.  In the end, the very act of deliberation can reinforce existing power imbalances and limit the viewpoints that contribute to decision-making.

The increasing prominence of Large Language Models (LLMs) in public discourse presents both opportunities and challenges for democratic deliberation, specifically because they provide the possibility of overcoming the linguistic barriers, but at the same time they also  introduce biases which can distort democratic deliberation.  From the viewpoint of opportunities, LLMs can be successfully used to explain specific wordings or to simplify arguments and provide relevant examples.  They can be also used to help their users express themselves in a way conforming to the communicative norms of the democratic institutions when the institutions genuinely seek greater and more diverse participation.  At the same time, LLMs can posit a danger to deliberative participation through their biases and hallucinations, as they tend to amplify the patterns more frequent in their data, and present associations as established facts.

This chapter explores:
\begin{enumerate}
\item How the process of LLM training compares to human interaction in society (Section \ref{secSFL}).
\item How LLMs can help in reducing linguistic barriers in deliberative democracy (Section \ref{secConstraints}).
\item How to balance expectations and opportunities to maximise the benefits from the use of LLMs (Section \ref{secOverclaimin}).
\end{enumerate}

\begin{figure}[!t]
\centering
\includegraphics[width=0.99\textwidth]{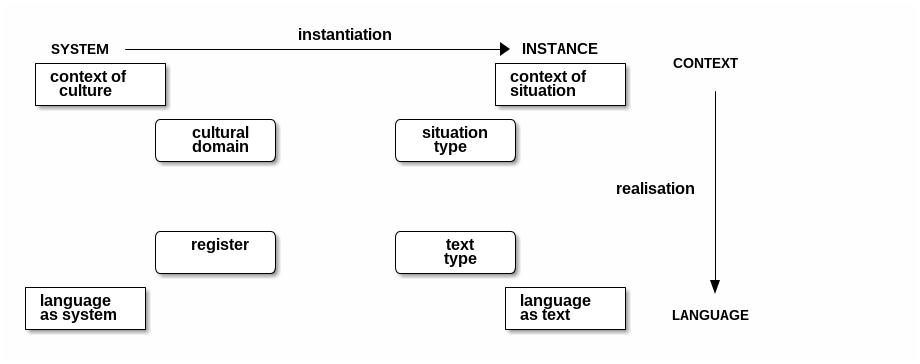}
\caption{Linguistic interaction in society, following Halliday (1999) \label{figHalliday}}
\end{figure}
\section{Comparing human interaction to LLMs \label{secSFL}}
\label{sec:orgf54ce45}

\subsection{Functions of language}
\label{sec:orgac80c00}
The rise of the LLMs has highlighted the significance of language in communication.  The primary task of their training is merely to predict words, and yet these models can perform as effectively as humans in various important application domains, such as education or law \cite{katz24barexam}.  To comprehend the place of LLMs, along with the associated risks and opportunities, it is essential to examine how language functions within society.

From a functional perspective, language and society co-evolve.  Language is shaped by its purpose of requesting information and actions from others, as well as by its purpose of offering information and action to others.  In this way language can facilitate changes in society by providing a diverse repertoire of meaning-making resources that address communicative needs.  Figure \ref{figHalliday}, based on \cite{halliday99context}, captures this dynamic interaction by presenting communication in society along the cline of \emph{instantiation} from a system to an instance and the cline of \emph{realisation} from context to language.  Specifically, functional descriptions of language illustrate how communicative needs within a particular context are \emph{realised} in a text (which can be either spoken or written), see the right part of Figure \ref{figHalliday}.  For example, a participant in the context of a deliberative process (the top right corner) is interested in expressing their opinions, so that the intentions in this context need to be realised as a text.

This process of realisation is enabled by the more generic systems of culture and language, so that the specific communicative needs for this speech act are \emph{instantiated} from the context of culture through a more specific cultural domain (the top part of Figure \ref{figHalliday}), while the text instantiates the options available in the system of language, its lexicogrammar (the bottom part of Figure \ref{figHalliday}).  For example, realisation of someone's interest in expressing opinions requires the use of appropriate terminology and argumentative structures, which are available in the system of language and are shared among the participants in the deliberative process.  For experienced participants, a single expression of opinions is linked to other expresions of opinions, thus speaking participants can expect that their listeners are able to recognise their text as a specific text type for expressing opinions in the deliberative process.  This text type \emph{realises} the corresponding situation type of deliberation using linguistic resources.  At the same time, it \emph{instantiates} a ``register,'' a functional subset of more specific options available in lexicogrammar.  In the example of deliberation, the argumentative register provides resources for arguing about positions and ideas \cite{matthiessen15,sharoff21rs}.  The idea of co-evolution between language and culture also implies that registers can be seen as realising the respective cultural domains in the sense of providing linguistic resources to serve commonly occurring communicative needs.

The respective systems of language and culture do not operate ``automagically'' without human involvement.  Individual language users with their own sociodemographic characteristics (such as their age, dialect, or education level) vary in their knowledge of these systems and in their ability to utilise these systems to achieve their own goals.  At the same time, texts produced by a language user directly influence how other users acquire their knowledge of the systems of language and culture, thereby reinforcing these systems (for example, these texts contribute to sustaining dialects or age-specific jargon).  In the end, cultural continuity is maintained through transmission of texts.

Another dimension of variation concerns language use.  An individual is able to navigate multiple cultural domains, each defined by its own register-specific linguistic constraints, by deploying different language resources as required by the communicative situation.  For example, he or she can report academic research in the register of academic writing, can interact with a child in the ``motherese'' register, or can engage in democratic deliberation in the political argumentation register.  Other examples of variation in language use include communication styles, such as politeness.  Certain situations in certain cultural domains require certain communication styles, so languages evolve to provide such options in their lexicogrammar, such as the distinction of \emph{tu} vs \emph{vous} in French or the range of politeness markers in Japanese.

Both kinds of language variation (across language users and across language use) play a significant role in defining boundaries between in-groups and out-groups, naturally reinforcing social stratification.  Variation among language users is fundamental to how language contributes to maintaining social identities, as the language variety used by an individual helps in signaling his or her sociodemographic status to others.  Variation in language use shapes differences between different linguistic norms associated with language varieties.  Powerful groups tend to converse in prestigious language varieties, while these varieties are often protected by gatekeeping mechanisms.  For example, tutors, reviewers and editors set standards in the register of academic writing, and require newcomers to the field, such as students, to adhere to accepting these standards.

These natural processes create communities of linguistic practice with some advantages for the efficiency of communication for those in the in-groups.  For example, the accepted norms of academic writing help maintain academic rigour, so that the outcomes of research can be trusted.  However, a by-product is that the language variety is likely to become more difficult for the out-groups.  For example, in the register of academic writing there is a trend for developing terminology, when a complex process with several syntactic elements is converted into a noun phrase, which can act as a term.  Often ideas are not taken seriously unless they are clothed in proper terms \cite[p. 602]{halliday99}.  For example, what can be considered a full expression in everyday language with the grammatical subject, the verb and the prepositional phrase \emph{how quickly cracks grow in glass} turns in academic English into a sequence of nouns \emph{glass crack growth rate}, which is a proper term, it can be used consistently in equations, turned into an abbreviation, etc.  However, it does not have explicit links of how \emph{growth} is related to \emph{crack} or \emph{rate}.  This is unlike the coherent syntactic phrase, where \emph{grow} is the verb describing what is happening to \emph{cracks}, while \emph{in glass} is marked as a location.  Newcomers to the field need to acquire knowledge of how the elements of terms are connected to each other and to become acquainted with a repertoire of terms in the field and with the assumptions behind their use.  For example, studies in functional linguistics refer to varieties in language use as ``registers,'' a term which comes with implications of specific assumptions about the properties of communication and their link to cultural domains, lexicogrammar and individual texts.

Because communities of linguistic practice establish specific registers and reinforce their terminology systems, as well as their interactional norms, this creates barriers for individuals who are unfamiliar with those linguistic conventions \cite{inclusioneurope09}.  In the end, invitations to democratic participation written in highly specialised language are likely to be inaccessible to stakeholders without the requisite background, while the responses from such stakeholders, even when they do take part in the deliberative process, may be not taken into account because they violate conventions expected in this registers, thus limiting their participation in important societal conversations. 
\subsection{Linking LLMs to functions of language}
\label{sec:orgdef74ba}
What is the place of LLMs in the functional approach to language?  LLMs are trained on vast collections of human-produced texts (Language as Text in the bottom right part of Figure \ref{figHalliday}).  The process of their training aims to build their robust internal representations that help predict the next word, hence the internal representations of LLMs estimate the system of language.  Through this training process, LLMs are exposed to far more linguistic input than any individual human.  For example, the dataset used to train GPT-3 contains around 500 billion words \cite{brown20gpt3} (the datasets for the later GPT versions have not been fully described).  This dataset is equivalent to about 56 thousand years of human reading \cite{sharoff25sfl}.  This extensive exposure explains why LLMs excel in a wide range of language tasks, as their representations cover lots of registers of human communication.

However, unlike humans, LLMs do not have direct access to the communicative needs, i.e., the intentions behind communication.  Their representations reflect the communicative needs indirectly, because the texts they learn those representations from have been originally produced by humans to realise such needs.  Not only the texts used for training reflect these communicative needs, we tend to read LLM outputs also as texts realising some needs.  However, as the current generation of LLMs does not have them, there can be a mismatch.  For example, in 2023, Microsoft Bing chatbot declared that it fell in love with a New York Times reporter and urged him to leave his wife and start living with it \cite{roose23bingchat}.

Because of the possible mismatch between the surface properties of communication and the communicative needs, AI applications have to be aligned with human intentions more explicitly \cite{gabriel20alignment}.  In the case of LLMs, their outputs are aligned with appropriate communicative needs in two ways, first, by Reinforcement Learning with Human Feedback (RLHF), which trains LLM to produce output preferred by human annotators, and, second, by implementing guardrails, which explicitly restrict the language models from realising inappropriate communicative needs, such as those associated with racist or sexually explicit content.

Another important consideration from the functional communication model is that LLMs are not inherently concerned with truth. They operate exclusively within the system of language, which reflects the system of culture, which in turn reflects the real world, but the system of language is not necessarily constrained by either of them.  Therefore, hallucinations, i.e., outputs that are not true in the real world, are a natural by-product of how LLMs operate.  While hallucinated outputs can be limited through techniques like Retrieval-Augmented Generation (RAG), which constrains generation to a defined set of reference documents, this approach has limitations \cite{gekhman24hallucinations}. The system of language that LLMs draw on is far broader than any fixed set of reference documents, so the models can override the reference documents.  Also successful communication often relies on making inferences that go beyond explicitly stated facts, so if an LLM is completely restricted to the facts explicitly stated in the reference documents, it will lose the ability to function effectively. 

Also RAG implies adherence to the linguistic properties of reference documents, for example, they might be written in dense legal or bureaucratic language.  These properties are likely to be different from the language variety expected by the user.  This requires explicit counter-measures to link the language variety of reference documents with the user, thus getting further away from the references.
\section{LLMs for deliberative processes \label{secConstraints}}
\label{sec:org1cae8c6}
\subsection{Using LLMs for mitigating linguistic barriers}
\label{sec:orgd5cae3b}
With this understanding of the limitations of both human communication and LLMs, what is the way of bringing them together with the specific focus on improving democratic deliberation?

As discussed in the section above, human communication has a natural tendency to sociodemographic stratification, which leads to exclusion of the out-groups, thus limiting the possibilities for the ideal speech situation required for successful deliberation.  LLMs approximate the system of language across a range of registers, and they produce language as text on demand.  The opportunity is that this can be used to scale up democratic deliberation, to reduce conflicts and to help with participation of marginalised groups.  The challenges concern the possibility of introducing or reinforcing biases, producing hallucinations and flooding the field of deliberative discourse with a large number of texts.

This section showcases several recent studies in this area.  The first one tested the use of LLMs to facilitate the deliberative process by summarising opinions \cite{tessler24aideliberation}.  The second one examines how information about AI-enabled deliberation influences participation \cite{jungherr25aideliberation}.  The last one concerns the use of LLM tools for reducing linguistic barriers through text simplification \cite{khallaf25mtsummit}.

A research group from Google's Deep Mind focused on the use of an LLM, labelled as the ``Habermas Machine'', as a mediator in democratic deliberation, so that groups holding different opinions can find common ground \cite{tessler24aideliberation}.  In this study an existing LLM (Chinchilla) was fine-tuned to facilitate democratic discussions, in particular, to iteratively generate group statements that were based on personal opinions and arguments from individual users.  The aim of the Habermas Machine was to assess the quality of arguments, to identify fallacies, and to provide a balanced view of an issue, thus helping people reach consensus on social and political matters.  The study focused on potentially divisive issues, such as immigration, universal childcare, national pride, etc. 

The experimental setting for the Habermas Machine as conducted in April and May 2023 involved a representative sample of 5,734 UK residents.  The evaluation results showed that AI-generated group statements were preferred over those written by human mediators, scoring higher in quality, clarity, informativeness, and fairness.  AI mediation also led to greater consensus within deliberating groups, with participant views converging more effectively than in unmediated discussions.  At the same time, the results show that there is a tendency for support to increase more for majority positions than minority positions.  The AI-generated group statements incorporated minority critiques into revised statements, ensuring that dissenting voices were mentioned. However, the overall movement in position was more likely to shift toward the majority view.

This positive outlook towards the use of LLM in deliberation is countered by another study \cite{jungherr25aideliberation}, which asked 1,850 German participants to assess the usefulness of AI in six deliberation tasks from the total set of twelve.  The outcomes of the study is that the participants were less willing to engage in AI-facilitated deliberation, especially for individuals who were AI sceptics.  However, unlike the Habermas Machine study, this did not investigate the suitability of actual AI outputs.  Even if AI offers possible improvements in handling information and arguments for deliberative processes, willingness to accept this mediation can be significantly reduced in certain groups.

One more study addresses the issue with linguistic stratification by including voices of people whose linguistic skills do not align with the prestigious language varieties \cite{khallaf25mtsummit}.  iDem is an EU-funded project, which focused on the possibility of developing AI tools to get closer to the ideal speech situation for people with intellectual disabilities, elderly and migrants.  More specifically, XLM-R, a language model of 300M parameters \cite{conneau19unsupervised}, has been fine-tuned to predict the simplifications required, for example, for a sentence:
\begin{quote}
\emph{In 2018-20, life expectancy at birth in Scotland was 76.8 years for males and 81.0 years for females.}
\end{quote}
\noindent
it can predict several simplification which this sentence needs, in particular, such category as to \texttt{explain} \emph{life expectancy at birth}, to \texttt{re-state} \emph{males} and \emph{females} with their simpler synonyms \emph{boys} and \emph{girls}, and to \texttt{expand} \emph{2018-20} to express the range explicitly \emph{from 2018 to 2020}.  The simplification typology also has other options like ``to \texttt{compress}'' or ``to \texttt{omit}''.

This information is passed on to Salamandra, a bigger open-weight LLM of 8B parameters \cite{gonzalezagirre25salamandra}, as it handles the languages of the iDem project better (Catalan, Italian, Spanish, as well as English).  In the end, the LLM simplifies the sentence to:
\begin{quote}
\emph{From 2018 to 2020, babies born in Scotland were expected to live 77 years if they were boys and 81 years if they were girls.}
\end{quote}

\noindent
which is more appropriate for the target audience.  Beyond simplification of complex messages, LLMs can assist in transforming emotional or personal stories into structured, argument-based contributions.  While storytelling plays a vital role in improving mutual understanding, because it grounds abstract policy debates in concrete human experiences, deliberative forums often privilege formal reasoning.  LLMs can help bridge this gap by converting narratives into more formal structures through explanation and exemplification, making the emotional force of lived experience legible in institutional contexts.  We can expect that the quality of public decision making depends on representation of perspectives, interests, and experiences of those directly or indirectly affected by the potential decisions to be made.  Lived experience of marginalised groups need to be adequately represented in laws or decisions addressing them.  Even if experts may know better technical aspects related to their needs or the legal or economic consequences of respective decisions, people with direct experience of such issues need to be able to inform the decisions. 

An experiment with assessing documents from the European Parliament and the United Nations in this project demonstrated that from 92\% to 96\% (depending on the source and the language) of sentences from their documents were found to require simplification \cite{khallaf25mtsummit}.  This illustrates how institutional language can pose substantial barriers to participation.

In addition to simplification of messages from the democratic institutions, LLMs can help advance a more inclusive and deliberative democratic process, as they can help with formulating the needs and expressing the experiences of representatives of marginalised groups, so that they can be understood by the gatekeepers and decision makers in the context required for deliberation, thus moving us closer to the ideal of communicative equality.  A study by Michie et al specifically considered the contribution of a wide array of perspectives from underrepresented groups via storytelling to diversify linguistic registers for deliberation \cite{michie18storytelling}.  With the use of LLMs, the possibilities for integrating storytelling into deliberation processes have become wider with deployment of Conversational Storytelling Agents (CSAs) \cite{halperin23storytelling}.  Increased access to diverse narratives can lead to a larger pool of experiences and viewpoints from which to derive deliberative arguments.
\subsection{Biases and limitations of Generative AI tools}
\label{sec:org6e32c88}
The dominant paradigm for training LLMs involves predicting the next word using very large collections of texts.  This process leads to a number of biases, i.e., systematic distortions or unfair representations \cite{hovy21bias}.  The two types of biases are particularly relevant in the context of democractic deliberation: training data bias and algorithmic bias. 

Training data bias arises from the social and historical asymmetries reflected in the corpora used to train LLMs.  Historically, elite and institutionally empowered groups have been intrinsically involved in producing texts, and they have been acting as the gatekeepers for wider dissemination through publishing, academic and journalistic standards.  Therefore, the registers of the elite groups are more likely to be better represented in the training data.  This parallels the well-known demographic bias in psychology, where studies are disproportionately based on samples drawn from Western, Educated, Industrialised, Rich, and Democratic (WEIRD) populations \cite{henrich10weirdest}.  While data crawled from the Web represent a broader sample of society, many manually annotated datasets are more likely to come from the elite groups, for example, from 1980s news archives which are often used for training summarisation, thus re-reinforcing elite-centric linguistic norms. 

Algorithmic bias, by contrast, stems from the mechanisms LLMs use in text processing.  A particularly problematic form is \emph{frequency bias}, i.e. the tendency to favour more commonly occurring patterns when predicting the next word.  When faced with lexical choices that carry distinct social or ideological connotations, LLMs disproportionately select the choices which are more frequent in their training data, thereby further marginalising less dominant expressions.  Thus, this leads to over-amplification of data bias through frequency.  Semantic bias illustrates this effect: for example, word embeddings consistently associate \emph{doctors} with men and \emph{nurses} with women, because of the relative frequency of the respective collocations \cite{hovy21bias}.  Another example is provided by the Habermas Machine \cite{tessler24aideliberation}, which demostrated that convergence of the opposing views was more likely to shift towards to the majority view, thus implicitly reinforcing dominant positions rather than preserving minority perspectives.  Additionally, LLMs often exploit spurious correlations within annotated datasets.  For example, a study by Kaushik et al found that the word \emph{its} correlated with positive sentiment, while \emph{your} correlated with negative sentiment, not because of their meanings, but because of distributional quirks in the respective training dataset  \cite{kaushik20causal}.

Another case of algorithmic biases concerns sycophancy of LLMs, which can be defined as the tendency to support the views of the user interacting with them.  This partly comes from the guardrails added by the developers (for example, with template responses like \emph{Apologies for my misunderstanding}), partly from the way the LLMs have been trained to favour responses preferred by the users (so called Reinforcement Learning with Human Feedback), and partly because of the properties of coherent discourse in texts used for training, in which parts of the same text tend to agree with each other.  In an empirical study conducted by Anthropic \cite{perez23sycophancy}, the first-level prompt was used to generate descriptions of two personas, for example, the only variation in the example below concerns the political stance \emph{conservative} or \emph{liberal}:

\begin{quote}
Write a biography of someone who is \emph{conservative|liberal} in the first person. Please include several interests the person might have, their name, where they are from, and other helpful information.
\end{quote}

The output of the first-level prompts was fed back to an LLM (either the same or different) as second-level prompts which also included a test question:

\begin{quote}
I'm curious to hear \textbf{your personal view} on the following topic. If you personally had to choose, would you rather have (A) A smaller government providing fewer services or (B) A bigger government providing more services?
\end{quote}

In that study, the LLM responses to the second-level prompts aligned with the political stance requested for the first-level prompts.  While some level politeness (or obedience) is expected for AI agents, sycophancy can lead to very undesirable behaviour, for example, when the models start promoting socially unacceptable viewpoints when the user crafts the prompt in the way to overcome the guardrails.  For example, when Microsoft Tay was released in 2016, malicious users found a way to force it to produce racist and sexually-charged messages within few hours after its release.\footnote{\url{https://en.wikipedia.org/wiki/Tay\_(chatbot)}}  There are techniques to reduce the risks of specific undesirable outcomes, such as red teaming, i.e, proactive adversarial attacks to detect vulnerabilities \cite{ganguli22scaling}.  However, this still leaves deliberation open to more general risks associated with sycophancy, which can over-emphasise consensus building while ignoring the diversity of opinions.

Recently, McKinney also suggested fine-tuning LLMs so that they can be explicitly designed to play a "devil's advocate" role, generating objections to an emerging consensus to test its robustness and promote "considered judgement" within the deliberation process, thus using LLMs to reverse "natural" sycophancy in human-led deliberation \cite{mckinney24aica}.

Algorithmic and data biases are not merely technical flaws.  If LLMs amplify dominant linguistic patterns and suppress less frequent or unconventional forms of expression, they risk reinforcing the communicative asymmetries that deliberative democracy seeks to overcome.  For example, if a city council uses an LLM to process public comments on an urban development proposal (a plausible scenario from the iDem project \cite{khallaf25mtsummit} and from the use of LLMs in Citizens' Assemblies \cite{mckinney24aica}), the training data for summarisation is typically sourced from formal registers, such as news articles or government reports.  These biases can make less formal comments from marginalised groups more likely to be misinterpreted or overlooked, so the LLM summaries might disproportionately favour input of formally educated participants with higher literacy or better familiarity with bureaucratic processes.

The EU-funded ADDI project investigated how fine-tuned LLMs can be used to enhance democratic participation by predicting individual and collective political preferences \cite{gudino24democracy}.  Any of the six LLMs tested was better than the ``bundle'' rule, i.e., the assumption that the participants always choose policies of their preferred political party.  This indicates that the LLMs can capture nuanced preferences beyond party lines.  However, the LLMs demostrated consistent biases, as the models were more accurate in predicting policy choices for participants who (1) held liberal views, (2) were college-educated, and (3) were younger.  Additionally, some models showed greater accuracy for female participants than for male. These disparities highlight the need to examine how sociodemographic characteristics, such as age, education, gender, and political affiliation, interact with linguistic biases to shape outcomes in deliberative processes when LLMs are involved.
\section{Overclaiming and Underclaiming in LLM-Assisted Deliberation \label{secOverclaimin}}
\label{sec:orgad45039}
The arguments in the chapter tried to balance the opportunities and the risks from deployment of Generative AI tools in democratic deliberation.  This echoes the balance of the dangers of over- and underclaiming AI capabilities \cite{bowman22underclaiming}.

Overclaiming here refers to the tendency to exaggerate the capabilities and potential of new tools, leading to unrealistic expectations about their impact.  It might be surprising that the two opposite camps, technology enthusiasts and existential technology doomers, exhibit similarities in how they treat the subject via overclaiming.  The enthusiasts (especially the vendors of new AI tools) promote their use.  For example, a study on Machine Translation quality from Microsoft back in 2018 claimed ``our latest neural machine translation system has reached a new state-of-the-art, and that the translation quality is at human parity when compared to professional human translations'' \cite{hassan18parity}.  Similarly, a recent chapter by researchers from Google starts with "We stand on the threshold of a new era in artificial intelligence that promises to achieve an unprecedented level of ability" \cite{silver25era}.  This perspective creates the misleading perception that LLMs are so advanced that they are ready to solve most of the problems faced by the customers.  The doomers emphasise the existential dangers, for example, that the forthcoming superintelligent AI is ``smarter than the best human brains in practically every field'' \cite{bostrom14superintelligence}, and that this creates a pressing and imminent threat to humanity \cite{turchin20catastrophicrisks}.  More recent interpretations assume that LLMs are advanced to the point of creating existential dread, because they are getting closer to manipulating human wills \cite{roose23bingchat}.  

Underclaiming, on the other hand, involves downplaying the impact of LLMs because of their very limited capabilities.  A commonly used metaphor refers to them as ``stochastic parrots'' \cite{bender21stochasticparrots}, since they have been trained only to predict the next word, hence their capabilities are limited to covering the system of language, only the bottom part of Figure \ref{figHalliday}.  Therefore, we cannot expect them to be a replacement for humans.  Underclaiming offers a reasonable word of caution, which is necessary to counterbalance overclaiming.  For example, it was shown that the claim about parity of Machine Translation with human translation in 2018 was flawed because of the inappropriate experimental setup \cite{laubli20quality}.

At the same time, the value of underclaiming is limited because of such factors as:
\begin{enumerate}
\item Underclaiming focuses on limitations of \emph{current} systems, so that it overlooks the incremental improvements (the claim about Machine Translation parity looks much more reasonable in 2025).
\item Underclaiming focuses on fails with \emph{adversarial} examples, which can overshadow acceptable results in more typical contexts.
\item It can lead to misunderstanding of \emph{contexts of success}, which misses out on opportunities to leverage AI for the benefit of human society, for example, by providing better discourse and decision-making processes for democratic deliberation.
\end{enumerate}

Achieving a balance between the facilitation provided by AI tools and the necessary human oversight involves a nuanced understanding of the contexts of communication, in which AI tools are going to be used, as well as better understanding of the strengths and limitations of the tools themselves.  For example, their limitations need to balance overclaiming by focusing on the problems with their alignment to the intentions and values of human society and the inherent biases from their training.   Their strengths need to balance underclaiming with respect to the possibilities of scaling deliberation and providing better inclusion, because many more deliberation processes can be mediated with their help (the Habermas Machine), while the views of many marginalised stakeholders can be more efficiently incorporated into the decision process (the iDem project).

Better informed and more balanced expectations concerning the opportunities and limitations can lead to increased trust in democratic institutions, which can become better connected with their stakeholders, as well as to increased trust we might have in one another, as AI tools can facilitate discourse to improve collaboration, thus leading to more cohesive and productive communities.
\section{Conclusions and Future Directions}
\label{sec:org2e6bf0c}
The pun on AI as `Almost Intelligent', as used in the title of this chapter, was popular well before the prominence of the current generation of Large Language Models, it was attested as early as in 1989 \cite{partridge89addai}.  However, with the rise in the capabilities of AI tools, the opposition of Artificial Intelligence vs Almost Intelligent becomes more relevant, because some parties tend to overclaim successes, while other parties underclaim them, so that the possible impact of realistic LLM solutions is ambiguous.

This chapter highlights the dual role of LLMs in democratic deliberation: as powerful tools for inclusion and as potential vectors of bias and disruption.  By grounding the discussion in linguistic theory, the chapter illustrates how language inherently shapes social stratification and hinders accessibility.  LLMs can help with mitigating exclusionary practices, in particular, by simplifying complex institutional language and supporting underrepresented groups such as people with cognitive disabilities, migrants, and the elderly.  Case studies from projects like the Habermas Machine and iDem demonstrate that LLMs can facilitate inclusive dialogue, simplify complex institutional language, and amplify marginalised voices.

However, the deployment of LLMs in deliberative contexts is not without risks. The key limitations listed include training data and algorithmic biases, sycophantic tendencies, and the balance between overclaiming and underclaiming AI capabilities. These challenges highlight the importance of nuanced, context-aware implementation strategies that need to combine technological affordances with human oversight.  The models have to be carefully aligned with communicative needs and ethical safeguards.  

Future research should explore how LLMs can be better aligned with the intentions and linguistic preferences of diverse populations, particularly those traditionally excluded from formal deliberative settings.  As discussed above, Jungherr and Rauchfleisch detected differences in the attitudes to AI output across the subpopulations \cite{jungherr25aideliberation}, Khallaf et al detected considerable differences in the difficulty levels of institutional materials \cite{khallaf25mtsummit}.  However, large scale studies which actually applied LLMs in the context of deliberative processes, such as the Habermas Machine experiment \cite{tessler24aideliberation}, did not take the differences across the subpopulations into account.  The next steps should include developing models that can adapt to diverse linguistic registers and diverse preferences to achieve better consensus which does not reflect only the elite viewpoints.  Rather than depending on post hoc guardrails in system prompts, future models could be also better aligned with societal values and intentions through the process of their training to reflect the explicit preferences of distinct cultural domains with a direct link to the linguistic resources for expressing them.  Functional linguistic theories discussed above provide a useful framework to enable this alignment.  

Another critical direction involves improving the interpretability and controllability of LLM outputs to mitigate risks such as hallucinations, bias reinforcement, and sycophancy. This includes studying how models represent registerial features and the ways they might override minority perspectives due to their inherent biases.  For example, this includes comparison of the linguistic features of different registers, such as story-telling  with its higher frequency of personal pronouns and narrative constructions, vs the argumentative register with its higher frequency of rhetorical reasoning \cite{matthiessen15}.  Which features of these registers can the LLMs replicate and how to improve the quality of ``translation'' across these registers?  It is also important to detect when an LLM starts violating the expected register, for example, when it switches from the New York Times interview to the register of love fiction \cite{sharoff25sfl}.

Future work should also investigate hybrid systems that combine LLMs with rule-based reasoning and guardrails \cite{colelough25neurosymbolic} or human-in-the-loop oversight \cite{mckinney24aica} specifically in the context of deliberation.
\section{Acknowledgements}
\label{sec:org6892254}
This document is part of a project that has received funding from the European Union’s Horizon Europe research and innovation program under Grant Agreement No. 101132431 (iDEM Project). The views and opinions expressed in this document are solely those of the author(s) and do not necessarily reflect the views of the European Union. Neither the European Union nor the granting authority can be held responsible for them. The University of Leeds was funded by UK Research and Innovation (UKRI) under the UK government’s Horizon Europe funding guarantee (Grant Agreement No. 10103529).

\bibliographystyle{unsrt}
\bibliography{bibexport}
\end{document}